\pdfoutput=1
\documentclass[10pt,letterpaper]{article}
\usepackage[T1]{fontenc}

\usepackage[pagenumbers]{westlakeagi}
\usepackage{tcolorbox}
\usepackage{fvextra}
\usepackage[pagebackref,breaklinks,colorlinks,allcolors=wagiblue]{hyperref}

\newcommand{\papertitle}{Code World Model: Coding Agent as World Brain}
\wagilogo{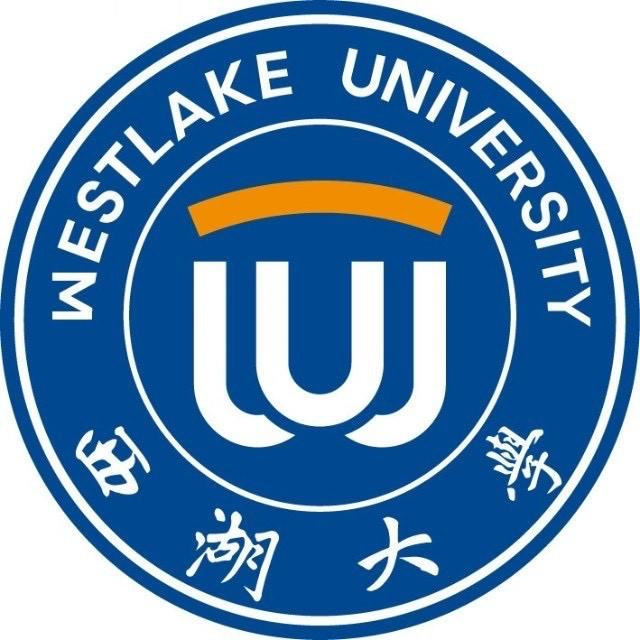}

\begin{document}

\vspace*{-0.42in}%
\wagibanner
\vskip 0.16in
\begin{tcolorbox}[colback=wagiblue!4, colframe=wagiblue!40,
                  boxrule=0.6pt, arc=10pt,
                  left=22pt, right=22pt, top=14pt, bottom=10pt]
  \begin{center}
    {\LARGE\bf \papertitle\par}
    \vskip 18pt
    {\large
     Yiwen Chen$^{1,2}$, \quad
     Guosheng Lin$^{2}$, \quad
     Chi Zhang$^{1}$\par}
    \vskip 6pt
    {$^{1}$AGI Lab, Westlake University \quad
     $^{2}$Nanyang Technological University\par}
    \vskip 4pt
    {\large\mdseries\ttfamily \href{https://buaacyw.github.io/cwm/}{\textcolor[HTML]{FF3B30}{https://buaacyw.github.io/cwm/}}\par}
  \end{center}
  \vskip 12pt
  \centerline{\large\bf Abstract}
  \vskip 7pt
  {\it\noindent World models aim to simulate how complex environments evolve under actions and events, yet existing video-based world models primarily learn dynamics from visual observations, which reveal outcomes rather than the underlying knowledge, rules, and mechanisms governing world evolution. This makes it difficult to maintain persistent consequences and support coherent, open-ended evolution. We introduce \textbf{Code World Model}, a framework that separates world evolution from visual realization by combining the reasoning and coding capabilities of language models with the generative priors of video models. A coding agent serves as the world brain, reasoning about events and their consequences and generating executable code to maintain persistent world state and perform rule-consistent evolution. To connect executable state with visual generation, we introduce a \textbf{proxy} representation that encodes frame-wise spatiotemporal constraints and is compiled into a proxy video, which conditions a video model to render high-fidelity visual observations. We further develop data pipelines for constructing aligned proxy--observation pairs from gameplay and real-world videos. After fine-tuning on paired gameplay data, MiniMax-H3 follows proxy-based spatiotemporal specifications from simple interactive worlds built by the coding agent while preserving rich visual details and dynamics. These results demonstrate the potential of combining code for persistent world evolution with video models for flexible visual realization, providing a new path toward open-ended world models.
\par}
\end{tcolorbox}
\vskip 0.35in

\begin{figure}[!h]
  \centering
  \includegraphics[width=\linewidth]{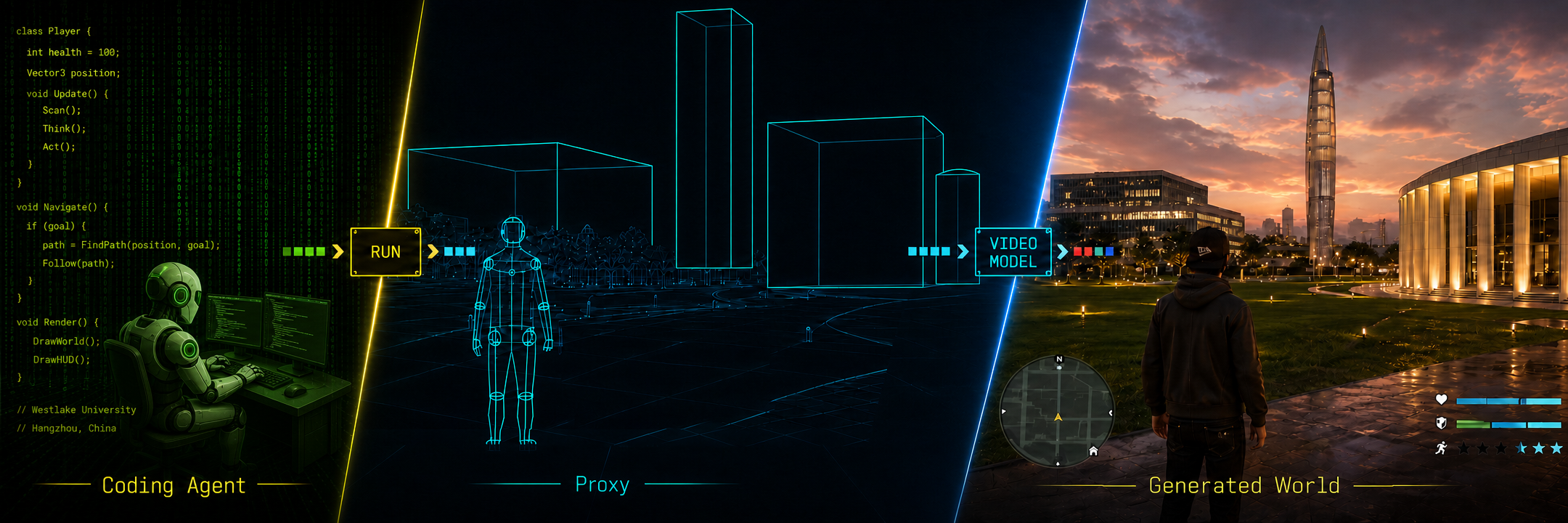}
  \caption{Overview of \textbf{Code World Model}: \textbf{Left:} a coding agent
  updates state via executable code; \textbf{Middle:} a lightweight compiler
  renders a coarse proxy video; \textbf{Right:} a video model generates
  high-fidelity observations from the proxy and text. This figure is
  AI-generated for illustration purposes.}
  \label{fig:teaser}
\end{figure}

\section{Introduction}
\label{sec:introduction}

A world model aims to represent the state of a world and how that state evolves in response to actions and events \citep{ha2018worldmodels,hafner2019learning,hafner2023dreamerv3}. Such models are essential for applications that must understand, predict, and act in evolving environments, including autonomous driving, embodied intelligence, and open-ended game worlds \citep{hu2023gaia1,yang2024unisim,bruce2024genie}. A complex interactive world may contain a large number of entities whose attributes, relationships, and behaviors evolve jointly over long time horizons. An action can affect not only the state of its immediate target, but also subsequent events and the future behavior of related entities. For example, in a game setting, actions by players, non-player characters, and environmental events can jointly reshape the world over time. If a player assassinates the ruler of a city, the local order, faction alliances, and the behavior of many non-player characters all change, and the consequences continue to unfold in subsequent interactions. Modeling such evolution requires broad world knowledge, an understanding of complex relationships among entities and events, the ability to infer consequences based on commonsense knowledge and world rules, and continual reasoning, planning, and decision-making. These capabilities are essential for building worlds that evolve coherently beyond predefined trajectories and support open-ended interaction; such worlds, in turn, provide a foundation for game generation and for training and evaluating agents \citep{yang2024unisim,alonso2024diamond,jang2025dreamgen,sharma2026worldgymnast}.

Among existing approaches to interactive world modeling, video world models have emerged as a particularly promising direction because they learn world dynamics directly from visual experience~\citep{kim2020gamegan,menapace2021playable,menapace2022playable_environments,micheli2023iris,hu2023gaia1,yang2024unisim,bruce2024genie,alonso2024diamond,valevski2024gamengen}. Recent systems primarily formulate the task as action- or prompt-conditioned video generation and have achieved substantial progress in visual quality and action controllability~\citep{menapace2024promptable,wu2024ivideogpt,xiang2024pandora,deepmind2024genie2,deepmind2025genie3,kanervisto2025wham,he2025video_simulators,gao2025adaworld,yu2025gamefactory,che2025gamegenx,li2025hunyuan_gamecraft,zhang2025matrix_game,xiang2025pan,huang2026vid2world,mao2026yume15,dreamx2026world,gao2026lingbot2}. Other efforts improve long-horizon temporal consistency and persistent memory~\citep{xiao2025worldmem,chen2025vrag,sun2025worldplay,hong2025relic,robbyant2026lingbot,wang2026matrix_game3,yu2025context_memory,li2025vmem,wang2026worlddirector,xiong2026actworld,huang2025self_forcing,wu2026infinite_world,yu2026mosaicmem}, while streaming and real-time systems support increasingly responsive interaction~\citep{decart2024oasis,reactor2026happyoyster,feng2024matrix,guo2025mineworld,he2025matrix_game2,cheng2025nfd,lin2025aapt,yang2025longlive,liu2025rolling_forcing,zhu2026causal_forcing,zhao2026minwm}. By learning from large-scale visual data, video models acquire rich priors over environments, agents, motion, and interaction, enabling highly expressive visual synthesis without manually crafting every appearance and motion pattern. Videos also naturally capture how complex worlds evolve over time.

Yet a central obstacle to building such world models is learning the hidden mechanisms that govern world evolution from visual experience alone. Videos record the observable outcomes of world dynamics, whereas the world knowledge, commonsense rules, inter-entity relations, long-horizon plans, and mechanisms of consequence propagation that produce these outcomes are not directly observable. This is particularly evident in gameplay videos: each frame is rendered by executing fixed game code, yet the executable logic that produces each frame is discarded after rendering and can be observed only indirectly through its sparse visual consequences. Moreover, many important state changes occur out of view or extend beyond the temporal context available to the model \citep{sun2025worldplay,yu2025context_memory,li2025vmem}. Scaling up training data provides more observed outcomes but no direct evidence of the underlying mechanisms: the model must still infer them from their visible effects. Although scaling up video training data can improve these capabilities to some extent, existing video world models have not yet demonstrated the commonsense reasoning, long-horizon planning, and continual prediction of consequences required for open-ended world evolution. Merely generating the next observation is therefore insufficient to sustain a complex interactive world as it continues to evolve.

Building on this motivation, a capable world model should combine two complementary strengths. First, it should continually maintain and operate a complete world, inferring the consequences of events from knowledge and world rules and allowing the resulting changes to persist throughout long-term evolution. Second, it should retain the ability of video models to generate rich visual appearance, motion, and fine-grained dynamics, rendering the evolving world into high-fidelity, open-ended visual observations. Achieving this goal requires several qualitatively different capabilities: broad knowledge and reasoning to decide how events should affect the world; the ability to maintain a persistent world state that records entities, relations, rules, and historical events, so that past changes continue to influence future evolution even when they are not currently visible; and generative visual capabilities that translate the evolved world into final observations. This naturally suggests a functional decomposition: world reasoning and execution determine how the state evolves, while visual generation renders the evolved state into visual observations. Crucially, these two sets of capabilities have already been extensively developed using different data sources, which motivates us to combine the knowledge and reasoning learned from text and code with the visual and motion priors learned from video.

The first strength calls for a component that can understand world knowledge, reason over events, and determine their long-horizon consequences. Language models are natural candidates for governing this evolution~\citep{vaswani2017attention,brown2020language,chowdhery2022palm,hoffmann2022training,ouyang2022training,openai2023gpt4,wei2022chain,yao2023react,schick2023toolformer,chen2021evaluating,yang2024sweagent,tang2024worldcoder,dainese2024gifmcts,curtis2025pomdp,lehrach2025code_world_models}. Unlike video models, which primarily learn dynamics from visual experience, language models acquire rich world knowledge from large-scale text and code and have demonstrated strong capabilities in understanding abstract concepts, modeling complex relationships, and performing long-horizon reasoning. They can leverage this knowledge to reason, plan, make decisions, and anticipate the potential consequences of events. A complex world, however, also involves a large number of fine-grained and repetitive state changes. Invoking a language model for every low-level interaction is feasible but computationally inefficient; moreover, critical outcomes often need to be rule-consistent and reproducible.

The second strength calls for an execution substrate that can realize these decisions through frequent, rule-consistent updates. Code provides a natural execution substrate for such updates. In a game, for example, a video model can render the visual appearance of an attack, but whether it hits should be determined by the current world state, including the distance to the target and the attack range. Without stable rules, players cannot form reliable expectations about how their actions affect the world. Meanwhile, recent agentic language models and coding agents have become increasingly capable of autonomously generating, modifying, and maintaining complex codebases \citep{kimi2026k3,deepseek2026v4,chi2026gamedevbench,jiang2026opengame}. Code therefore need not be limited to static scripts predefined by developers: a coding agent can invoke, compose, and revise executable mechanisms as the world state, goals, and interactions change.

\begin{figure*}[t]
  \centering
  \includegraphics[width=\textwidth]{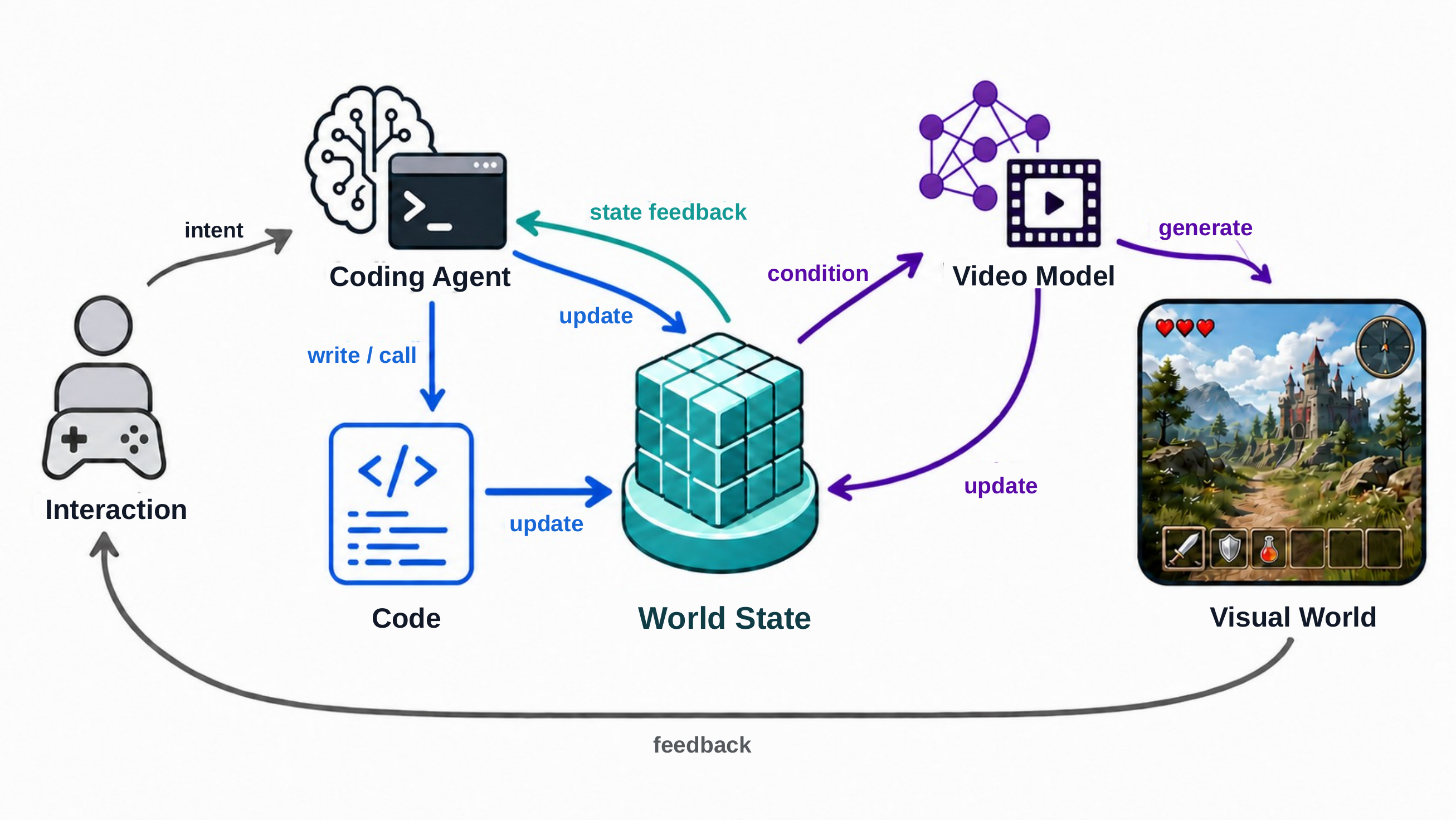}
  \caption{\textbf{Pipeline of the proposed Code World Model.} The coding agent serves as the brain of the world model, translating interaction intent into code that updates the world state. The updated world state is compiled into a proxy and paired with a text prompt to condition the video model, which generates the corresponding visual observations. Generated observations and state feedback then support subsequent interaction and world evolution.}
  \label{fig:pipeline}
\end{figure*}

Motivated by these observations, we propose \textbf{Code World Model} (Fig.~\ref{fig:teaser}), a world-modeling framework that comprises an executable world-evolution mechanism driven by a coding agent, a programmable interface between world state and visual generation, and a video model that produces the final visual observations. At its core, the coding agent serves as the world brain: it governs world state and its evolution through code, while the video model renders the evolved world into visual observations. Knowledge and rules thus drive changes in world state, while the world's visual realization draws on learned priors, allowing an interactive world to evolve continually and coherently beyond predefined trajectories. Fig.~\ref{fig:pipeline} summarizes how the coding agent updates world state through code and how the resulting proxy and prompt condition the video model to generate visual observations.

The coding agent and code operate in two distinct computational regimes. The coding agent handles infrequent but complex decisions, such as interpreting new events, reasoning about long-term consequences, selecting mechanisms, and revising how the world operates; it translates these decisions into executable, reusable, and modifiable code. Code then performs high-frequency, repetitive, and deterministic operations---updating positions and numerical attributes, resolving collisions, and executing rules---without invoking the coding agent at every step. This agent--code organization decouples the frequency of high-level reasoning from that of low-level state updates, allowing world state to evolve continuously while its operating mechanisms remain modifiable.

With this executable world state in place, a central technical challenge is to translate it into a conditioning representation that a video model can reliably interpret. One straightforward solution is to express the state as structured text; however, current video models cannot yet reliably translate dense, constantly changing textual specifications into observations with precise frame-wise positions, spatial relations, and camera trajectories. An alternative is for the coding agent to construct a complete, explicit 3D world and condition the video model on its renderings \citep{hunyuanworld2025,wang2026worldgen}. Although this provides stronger spatial control, constructing production-quality geometry, assets, animation, materials, and rendering systems is costly and prematurely constrains the final visual realization. To retain the flexibility of text-based specifications while recovering the spatial control of explicit 3D representations, we introduce \textbf{proxy}, a coarse representation of world state that encodes entity positions, poses, trajectories, spatial relations, and camera motion. A lightweight deterministic compiler renders it into a \textbf{proxy video}, which provides the video model with direct spatiotemporal constraints while leaving appearance and fine-grained dynamics unspecified. The proxy need only specify coarse spatial relations and overall trajectories, while the video model synthesizes local motion and physical interactions from its learned priors. Conditioned jointly on the proxy video and structured text, the video model generates the final observation using its learned priors over appearance, motion, and interaction. In this way, code determines what happens in the world and which consequences persist, while the video model determines how these outcomes are visually realized. Proxy therefore extends the coding agent's control from language into visual space, allowing it to express evolving world state as adjustable frame-wise constraints for the video model.

To instantiate and evaluate the interface from executable world state to visual observations, we develop a data-construction scheme for both gameplay and real-world videos. Our current prototype first adapts a pretrained video model using gameplay data. During gameplay, the target video and the camera, entity, and scene states required to construct the proxy are recorded synchronously from the same execution, yielding spatiotemporally aligned proxy--observation pairs. Real-world videos can likewise be paired with aligned proxies constructed from 3D reconstructions and object annotations, without requiring action labels and thereby substantially lowering the annotation burden. Fine-tuned on our paired gameplay data, MiniMax-H3~\citep{minimax2026h3} qualitatively adheres to the proxy specifications for character locations, action trajectories, scene layouts, and camera motion when conditioned on proxies recorded from simple interactive worlds built by the coding agent, while preserving high-fidelity appearance and fine-grained dynamics. Compared with control interfaces that express only coarse intent through text or low-dimensional signals for action and camera control \citep{sun2025worldplay,wang2026matrix_game3,gao2026lingbot2}, our interface directly specifies frame-wise entity motion and viewpoint changes, enabling finer-grained and more direct spatiotemporal control.

Overall, Code World Model unifies knowledge-driven, rule-consistent world evolution with the visual, motion, and interaction priors learned by video models. Our current prototype provides preliminary evidence for the viability of the framework's two key components: the coding agent can modify the state and operating mechanisms of a world by writing and revising code, and the fine-tuned video model can render the world state expressed by the proxy into corresponding visual observations.

Our contributions are summarized as follows:

\begin{itemize}
\item We introduce \textbf{Code World Model}, a framework in which a coding agent governs long-horizon world evolution through executable code, while a video model renders the evolving world state into high-fidelity visual observations.

\item We design a proxy-based interface that translates executable world state into a proxy video, providing direct frame-wise spatiotemporal constraints without requiring a complete high-quality 3D asset and rendering pipeline.

\item We construct a spatiotemporally aligned gameplay data pipeline and demonstrate geometry-assisted offline proxy construction on real-world recordings, providing preliminary evidence that proxy can serve as a learnable interface to world state.
\end{itemize}

\section{Related Work}
\label{sec:related-work}

\subsection{Interactive video world models}

Scalable diffusion transformers underpin many recent high-fidelity video generators, including HunyuanVideo and Wan \citep{peebles2023dit,kong2024hunyuanvideo,wan2025video}. Learned interactive environments further formulate world evolution as conditional prediction of future visual observations. Early work learns action-conditioned visual dynamics from gameplay or infers latent action spaces from unlabeled video~\citep{kim2020gamegan,menapace2021playable,menapace2022playable_environments,micheli2023iris}, while diffusion-based systems demonstrate that generative models can serve as playable neural simulators~\citep{bruce2024genie,alonso2024diamond,valevski2024gamengen}. Recent systems scale this paradigm toward language-guided simulation, action-controllable environments, and real-time navigable worlds~\citep{menapace2024promptable,wu2024ivideogpt,xiang2024pandora,deepmind2024genie2,deepmind2025genie3,reactor2026happyoyster}. Subsequent studies extend this formulation toward online control, streaming generation, and longer interaction in game environments~\citep{decart2024oasis,feng2024matrix,guo2025mineworld,he2025matrix_game2,kanervisto2025wham,he2025video_simulators,gao2025adaworld,huang2026vid2world}. A parallel direction broadens the generated scenes and control interface through open-domain game transfer, free-form actions, keyboard and mouse inputs, and text-directed world events~\citep{yu2025gamefactory,che2025gamegenx,li2025hunyuan_gamecraft,zhang2025matrix_game,xiang2025pan,mao2026yume15}. Together, these works establish action-conditioned video generation as a direct route to visually rich interactive rollouts without first constructing a complete explicit 3D world.

Long-horizon generation introduces a second set of challenges: errors accumulate autoregressively, while scene information outside the current context becomes difficult to recover. Recent systems address these problems through geometry-aware history, longer-context training, and broader general-domain data~\citep{xiao2025worldmem,chen2025vrag,sun2025worldplay,hong2025relic,robbyant2026lingbot,wang2026matrix_game3,dreamx2026world,wu2026infinite_world,yu2026mosaicmem}. Others introduce retrieved historical frames, persistent view representations, or event- and action-aware memory to preserve information that remains relevant later in a rollout~\citep{yu2025context_memory,li2025vmem,wang2026worlddirector,xiong2026actworld}. Complementary work reduces train--inference mismatch and deployment cost through self-rollout training, efficient long-video generation, causal distillation, and full-stack inference systems~\citep{huang2025self_forcing,cheng2025nfd,lin2025aapt,yang2025longlive,liu2025rolling_forcing,zhu2026causal_forcing,zhao2026minwm}. Most closely related to our setting, LingBot-World~2.0 augments a causal video generator with pilot and director agents, enabling semantic actions and mid-rollout world intervention~\citep{gao2026lingbot2}.

These advances progressively strengthen control, memory, and agentic interaction around a learned visual rollout. Their central modeling object, however, remains the generated observation sequence: scene-specific state and interaction outcomes are primarily carried by visual history, latent context, or visual memory. Code World Model instead places a coding agent and its persistent code in charge of an explicit world state, including the consequences that must continue to affect later interaction. The video model remains responsible for generating the world's appearance and dynamics, but receives the relevant state through the proxy rather than being required to internalize the entire process of world evolution.

\subsection{Generative 3D Worlds}

Generative 3D world models pursue explicit scene representations that support navigation, spatial editing, and free-viewpoint rendering. One line expands generated images or panoramic observations into connected meshes, Gaussian representations, or other explorable 3D scenes \citep{hollein2023text2room,yu2024wonderworld,lu2024genex,hunyuanworld2025}. Another combines video or multi-view generation with reconstruction to increase navigable scene coverage and accelerate 3D world creation \citep{schneider2025worldexplorer,yang2025matrix3d,li2025flashworld}. More recent systems emphasize extendable scene generation, language-guided asset and layout composition, traversability, and unified world generation and reconstruction \citep{li2025worldgrow,bian2025holodeck2,wang2026worldgen,hyworld2026}. These directions demonstrate the value of making geometry explicit when exploration, viewpoint consistency, or downstream graphics integration is required.

In these systems, the generated geometry, assets, and rendering pipeline constitute the visual world from which observations are produced. Our proxy serves a different purpose. It is not intended to be the final 3D world or to prescribe production-quality geometry and appearance; it is a coarse, programmatically constructed condition that communicates camera motion, entity layout, trajectories, and interaction-relevant state to the video model. The video model then generates the final appearance and fine-grained dynamics under these constraints. Simple 3D primitives may therefore be used to express the proxy, but explicit 3D construction is an interface to visual generation rather than its endpoint.

\subsection{Coding agents and world models}

Programs provide another representation of environment dynamics. WorldCoder \citep{tang2024worldcoder} has an LLM agent build a world model by writing a Python program and refining it through environment interaction, and \citet{curtis2025pomdp} induce probabilistic programs for POMDP model estimation under LLM guidance. \citet{dainese2024gifmcts} generate Python transition models with an LLM guided by Monte Carlo tree search for model-based reinforcement learning. \citet{lehrach2025code_world_models} encode state transitions, legal actions, observations, rewards, and termination as program functions consumed by a planner for general game playing; follow-up work distills this synthesis into lightweight models \citep{serapio2026gamecwm} and applies coding agents to build executable world models for interactive benchmarks \citep{rodionov2026executable_world_models}. This line shows that world rules and transition structure can be externalized as code that is inspectable, testable, and revisable. \citet{fair2025cwm} instead mid-train an open-weights LLM on Python execution traces and agentic interactions so that the model learns to simulate code execution, using world modeling to strengthen code generation and reasoning.

Recent agentic language models report strong performance on long-horizon coding and software-engineering tasks \citep{kimi2026k3,deepseek2026v4}. A complementary direction studies coding agents as builders of playable software, including benchmarks for multimodal game-development tasks and agentic systems for end-to-end game creation \citep{chi2026gamedevbench,jiang2026opengame}. These works demonstrate the growing ability of coding agents to construct and modify complex interactive programs, but focus primarily on producing the game artifact rather than using persistent, revisable code as the operating medium of a learned visual world model. In Code World Model, the coding agent instead governs the evolution of an open visual world through persistent code, and a learned video model converts state-derived proxy conditions into high-fidelity visual observations.

\section{Method}
\label{sec:method}

Our goal is to build open worlds that approach the causal complexity of the real world while maintaining high visual fidelity. We believe that building such worlds requires a \emph{coding agent} to serve as the world brain, \emph{code} to efficiently execute continuous, high-frequency state updates, and a \emph{video model} to generate high-fidelity visual observations with realistic motion. To achieve this goal, we introduce \textbf{Code World Model}, which unifies these components within a single framework. As summarized in Fig.~\ref{fig:pipeline}, the coding agent updates world state through code, and the resulting proxy and prompt condition the video model to produce visual observations.

Section~\ref{sec:cwm} defines Code World Model and explains how a coding agent governs world evolution through continuously maintained, executable, and modifiable code. Section~\ref{sec:conditioning} studies how world state can be communicated to the video model. It begins with structured text and analyzes its practical limitations. To tackle this challenge, we introduce a coarse visual condition, termed the proxy, that is deterministically generated from world state and provides the video model with direct spatiotemporal constraints. Building on this proxy design, Section~\ref{sec:data} presents a data pipeline for training the video model to understand and follow proxy conditions.

\subsection{Code World Model}
\label{sec:cwm}

To make the challenges posed by such complex interactions concrete, we return to the fantasy-world example from the Introduction. After a player assassinates a city's ruler and leaves, that event should continue to affect succession, local order, trade, faction relations, and the beliefs, goals, and behavior of many non-player characters. These consequences may keep developing while the city is off-screen, interact with later events, and become visible only when the player returns much later. A system must therefore determine not only how the city should look upon return, but also what happened during the player's absence, why it happened, which consequences remain active, and how the current situation will shape future evolution.

From the perspective of running the world, such complex interactions present two different challenges. The first is sparse but semantically complex reasoning: interpreting new events, connecting them to world knowledge, character relations, and social structures, determining which entities and mechanisms are affected, and planning consequences over long periods. The second is dense and repetitive execution: maintaining positions, attributes, schedules, cooldowns, collisions, and numerical rules, and translating high-level decisions into concrete state updates. A complex world requires both. The latter must operate at a substantially higher update frequency than the former, whereas the former demands substantially more sophisticated reasoning.

Video models provide valuable learned priors for how environments, characters, and interactions evolve visually \citep{yang2024unisim,bruce2024genie,alonso2024diamond,valevski2024gamengen}. Theoretically, sufficiently capable video models could acquire the capabilities needed to address both challenges from visual experience. In practice, however, video data reveal world rules only indirectly through their outcomes; off-screen character states have no continuous visual record; and consequences that propagate across locations and long time spans are sparse \citep{yu2025context_memory,li2025vmem}. The training context windows of current video models are typically shorter than one minute, whereas many causal processes that require reasoning unfold over days or even years in world time. This makes it particularly difficult for video models to acquire the capabilities needed to address the first challenge: the semantically complex reasoning required to construct and evolve a coherent world. Moreover, current video world models are trained primarily on gameplay videos, which preserve only the visual outputs of running fixed game code. The model must therefore infer the underlying rules and state transitions from those outputs alone, making this learning process highly inefficient and ultimately suboptimal.

The world knowledge, relational reasoning, and long-horizon planning capabilities of language models make them a natural choice for governing world evolution \citep{tang2024worldcoder,dainese2024gifmcts,curtis2025pomdp,lehrach2025code_world_models}. An agent can organize these capabilities into a continuing loop: read the current world, decide what should happen next, observe the result, and revise subsequent decisions. The agent should not, however, become the step-by-step executor of the entire world. As a world grows, thousands of world-state variables may require frequent updates, covering characters, objects, and global processes such as weather, traffic, and scheduled events. Invoking a language model for every such update would incur high inference costs and fail to sustain the update frequency and responsiveness required for real-time operation. A more scalable organization converts high-level decisions into reusable code and lets that code execute across entities, time, and events.

This organization aligns naturally with the central strength of contemporary coding agents: translating high-level intent into code that can be executed, reused, and continuously revised \citep{chi2026gamedevbench,jiang2026opengame}. Suppose that, after the ruler is assassinated, the coding agent decides that the city should enter an emergency state. It may change faction goals, activate existing behavioral rules, or add new mechanisms for patrols, succession, and message propagation. The coding agent need only decide the high-level behavior of each character and intervene when a goal changes, an exception occurs, or a mechanism must be revised. At other times, code continuously updates positions, orientations, and other runtime state at high frequency according to existing rules. Its execution rate is decoupled from both the coding agent's reasoning frequency and the video model's generation frame rate. In a large battle, code can continuously process distances, collisions, attack ranges, cooldowns, hits, damage, and chained triggers for many entities, while the coding agent focuses on sparse high-level intent, exceptions, and mechanisms that must be added or changed.

The result is a continuing agent-code loop. An interaction or world event first changes the current situation. The coding agent reads the maintained world state and chooses either to invoke existing code or locally modify the world program. Code then executes reusable mechanisms and advances many state variables. Execution results, tests, and new world feedback return to the coding agent as evidence for its next decision. Most repetitive updates require no additional model call, and world state may advance in an event-driven manner or at a frequency higher than visual generation. When existing mechanisms cannot handle a new situation, the coding agent can adaptively revise the world program, changing not only the current state but also how the world will operate in the future. Code is therefore neither permanently frozen game logic nor a controller independent of the agent; it is the coding agent's continuously maintained, executable, and modifiable extension.

Building on this insight, we introduce Code World Model: a world model that uses a coding agent as its brain. The core of Code World Model is to make the coding agent the primary intelligence governing world evolution. Through executable and modifiable code, it controls world state, operating mechanisms, and subsequent evolution, while code performs dense and reusable low-level updates on its behalf. This organization distinguishes Code World Model from existing video world models, which primarily govern generation through action, camera, or semantic conditions supplied to video model rollouts \citep{sun2025worldplay,wang2026matrix_game3,dreamx2026world,gao2026lingbot2}. Rather than merely supplying high-level guidance, the coding agent directly participates in the executable evolution of the world.

\paragraph{From world state to visual observations.}
The agent-code loop determines how a world operates, but it does not determine how that world should be visually instantiated. Because the paradigm is defined by how the world is governed and evolved, its visual backend is selectable. A seemingly direct solution is to have the coding agent construct an explicit 3D environment \citep{hunyuanworld2025,wang2026worldgen}. This route is effective when the required entities, appearances, actions, and interactions are known in advance, because each can be supported by prepared assets, animations, and simulation rules. An open world, however, can continually introduce new characters, objects, behaviors, and situations that were not anticipated by the existing pipeline. Supporting them requires not only new executable logic, but also corresponding visual assets, motion, and physical responses. More fundamentally, the ceiling of this route is set by the fidelity and coverage of its assets and simulators, making it difficult to approach the visual richness and physical complexity of the real world. This dependence makes explicit 3D difficult to scale toward open-ended visual worlds.

A video model can instead map conditions supplied by the executable world directly to visual observations without requiring a complete asset-animation-simulation-rendering chain. Learned from large-scale visual data, modern diffusion-transformer video models provide rich priors over appearance, motion, interaction, and physical behavior \citep{peebles2023dit,kong2024hunyuanvideo,wan2025video}. Code determines what occurs in the world and which outcomes affect subsequent interaction; the video model generates how those outcomes unfold and appear. Compared with a complete explicit 3D realization, this route offers a higher ceiling for visual realism and fine-grained interaction and can absorb broader real-world motion and physical priors as data, models, and compute scale, without expanding each asset, animation, and simulation component separately. This scalability supports both high-fidelity virtual worlds and visual environments closer to the real distribution required for VLM and embodied agent training. We therefore use a video model as the visual backend of Code World Model.

The primary cost of this choice is that every new visual observation requires generative inference, leading to higher marginal computation and latency than conventional rendering. For our objective of maximizing visual fidelity and open-ended generation, we consider the quality gain worth this cost. We expect video model inference to become increasingly efficient, making this trade-off more favorable over time \citep{yang2025longlive,zhu2026causal_forcing,zhao2026minwm}.

\paragraph{Formulation.}
Let $S_t$ denote the complete world state at time $t$, $A_t$ an action applied to the world by a player, the environment, or another agent, and $O_t$ the visual observation received by a user or downstream agent. Under the conventional formulation of a world model, the current state and action determine the next state, which subsequently produces the next observation \citep{ha2018worldmodels,hafner2019learning}:
\begin{equation}
S_{t+1}\sim p(S_{t+1}\mid S_t,A_t),
\qquad
O_{t+1}\sim p(O_{t+1}\mid S_{t+1}).
\label{eq:wm}
\end{equation}
In Code World Model, the complete world state must support both executable world evolution and visually coherent generation. We therefore decompose $S_t$ into an executable state $S_t^{\mathrm{exe}}$ and a visual state $S_t^{\mathrm{vis}}$:
\begin{equation}
S_t=\left(S_t^{\mathrm{exe}},S_t^{\mathrm{vis}}\right),
\label{eq:state-decomposition}
\end{equation}
The executable state $S_t^{\mathrm{exe}}$ contains the evolving world program, entity attributes, rules, relations, event history, and other variables that can be directly operated on. The visual state $S_t^{\mathrm{vis}}$ contains appearance, motion, and other visual information produced by the video model that must remain consistent over time.

These two parts are updated through different but coupled processes. We use $\mathcal{T}_{\mathrm{AC}}$ to denote the joint agent-code transition. Code repeatedly executes high-frequency deterministic updates under existing rules, while the coding agent invokes, composes, or revises these mechanisms according to its goals and feedback from the world. Together, they produce the next executable state. We use $G_\theta$ to denote the video model, which generates the next visual state from the updated executable state and the previous visual state. Their updates are written as
\begin{align}
S_{t+1}^{\mathrm{exe}}
&=\mathcal{T}_{\mathrm{AC}}\left(S_t^{\mathrm{exe}},A_t\right), \\
S_{t+1}^{\mathrm{vis}}
&\sim G_{\theta}\left(S_t^{\mathrm{vis}},S_{t+1}^{\mathrm{exe}}\right).
\label{eq:cwm}
\end{align}
The resulting visual state determines the observation presented to the user or downstream agent. Visual information that must persist is retained in world state, managed by the coding agent, and provided again when the relevant character, location, or event reappears. Writing $S_{t+1}^{\mathrm{exe}}$ as an input to $G_\theta$ expresses only an abstract conditional dependency; it does not mean that raw code or the complete executable state is passed directly to the network. Section~\ref{sec:conditioning} explains how the relevant information is selected and represented as a condition for the video model.

\subsection{World-State Conditioning for the Video Model}
\label{sec:conditioning}

Once a video model is selected as the visual backend, a direct question follows: how should the coding agent's high-level intent and the continuously changing world state produced by code execution be converted into a condition that the video model can use effectively?

The most direct solution is to organize executable state as structured text, since recent interactive video models already accept language conditions \citep{menapace2024promptable,xiang2024pandora,mao2026yume15}. The coding agent may write semantic descriptions of appearance, role, and goals for every persistent identity and store them in identity records managed by code. As the world runs, code continuously updates entity position, orientation, relations, behavior, and combat state. The system can automatically organize these persistent descriptions and runtime variables by identity, time, and event; merge them with the coding agent's current high-level intent and original text prompt; and provide the result to the video model. This route introduces no new condition modality and naturally accommodates new objects, attributes, and rules. It is therefore the simplest, most elegant, and apparently most scalable choice.

However, our experiments with recent video world models trained at scale \citep{gao2026lingbot2} show that text conditioning, even when paired with a dedicated camera-conditioning pathway, still fails to provide precise control over camera trajectories and entity motion. This observation does not imply that language cannot express precise world state. Given a sufficiently long description, language could theoretically specify every pixel of every frame. However, such a description would be extremely token-inefficient and would make it difficult to meet the latency requirements of real-time interaction. Current video models also struggle to follow instructions of such complexity reliably.

\paragraph{Proxy.} These experiments show that a primary challenge is enabling the video model to follow precise, high-frequency updates to entity positions, spatial relations, and camera trajectories. This motivates us to explore a different conditioning mechanism that provides more direct and fine-grained spatiotemporal control.

Since the target output is itself a video, a natural choice is to express this control in the same spatiotemporal modality. Such a condition can specify camera motion, entity location, occlusion, relative relations, and trajectories directly in image coordinates. The video model then reads control information whose spatiotemporal organization is already established. Compared with text alone, this provides clearer grounding in a representation that the model can more readily interpret.

The information in this visual condition should come entirely from world state jointly constructed and executed by the coding agent and code. It should not be supplemented by another information source or generative model, such as a learned 3D generator, because that would introduce a mapping that the coding agent cannot directly inspect or control. Nor should the coding agent be required to produce a complex visual world: doing so would raise construction cost and prematurely prescribe the final image. Every control signal in the resulting condition should be traceable to world state, keeping the state-to-condition path white-box to the coding agent.

We organize the camera, entity positions, poses, trajectories, and spatial relations that affect the current observation into a coarse programmable world representation, which we call the \textbf{proxy}. A deterministic rendering process converts the proxy into a spatiotemporal visual condition, which we call the \textbf{proxy video}. Structured text and the proxy originate from the same world state but use different condition representations. Text serializes selected state as language; the proxy organizes it spatially before it is supplied to the video model as a proxy video. In this way, the coding agent controls the video model through two complementary channels: text communicates semantic intent, while proxy video expresses evolving world state as adjustable frame-wise constraints.

The proxy is not intended to represent the target world as completely as possible. It retains only the coarse structure and interaction state that the current observation must obey. It does not attempt to express textures, materials, fine lighting, or other appearance information that the video model should generate. Its goal is not to provide a low-quality target video, but to express required state and spatiotemporal relationships in the simplest useful form.

Constructing the proxy and rendering the corresponding proxy video also does not require a complete 3D asset-production or high-quality rendering pipeline. The coding agent programmatically calls and combines a given set of simple reusable primitives. Each primitive can be expressed by only a few lines of code and therefore adds little representational burden for the agent. Addressable parameters specify entity position, scale, pose, layout, camera, and state markers; a simple deterministic compiler then performs lightweight rasterization. The process requires no production-quality assets, complex materials, lighting, fine animation, or expensive rendering. The proxy is therefore constructible, inspectable, addressable, and locally editable rather than an opaque visual proposal from a learned 3D generator. Although the final condition enters the video model as image or video tokens, its control information still originates from world state.

The proxy is used jointly with structured text. It expresses coarse spatiotemporal state that the current observation must obey, while text specifies character and object appearance, action semantics, and dynamic details left unspecified by the proxy. Adding a proxy requires the coding agent to construct a coarse spatial representation, but in return it provides clearer spatiotemporal grounding. Text alone requires no visual representation and leaves the largest space for the video model's realization, but makes precise spatial control more difficult.

\paragraph{Condition bandwidth.}
The central proxy-design choice is its \emph{condition bandwidth}. A richer proxy provides stronger grounding of entities, layout, and camera, but asks the coding agent to construct and maintain more primitives and parameters. For example, if the proxy encodes a character's articulated motion at the joint level, the coding agent must also control that character's joint trajectories during inference. Current coding agents still struggle to implement such joint-level motion control reliably, which can limit final performance. A proxy with too little information is lighter to construct but may fail to impose sufficient spatiotemporal constraints. We formulate proxy design as a trade-off between constructability and grounding strength: it should provide the minimum sufficient state that the current observation must obey, while leaving final appearance and fine-grained dynamics to the video model.

\noindent\textbf{Discussion.} A natural concern is that a visual condition may introduce too many tokens. In our implementation, the proxy uses one-quarter of the target video's resolution along each spatial dimension and therefore adds only $1/16$ as many visual tokens as the target video to the omni transformer, making its additional inference burden negligible.

The proxy should not be interpreted as a globally fixed condition. Instead, it is a visual prompting tool through which the coding agent controls the video model, serving as the visual counterpart of a text prompt. In a complete Code World Model, the coding agent itself determines whether to enable, disable, or adjust the proxy according to the current world and generation task. When text alone is sufficient, it can omit the proxy; when an observation must closely follow positions, trajectories, spatial relations, or interaction topology, it can activate the relevant proxy. The coding agent can further select interaction-critical entities and regions and adjust the represented state types, spatial granularity, resolution, and condition frame rate. As video model control improves, the proxy may become sparser or disappear in selected situations. This paper instantiates and evaluates one fixed proxy design; autonomous switching and bandwidth adjustment remain part of the broader paradigm's design space.

The discussion above treats the language model and video model as separate components, but the condition-design problem is independent of that implementation choice. Even if they are unified into an omni-modal network, continuously changing external world state maintained by code must still be encoded and supplied to the visual generation process. Parameter sharing may simplify communication between components, but it does not remove the state-to-condition problem.

\subsection{Data}
\label{sec:data}

Building on the proxy design above, we require paired proxy and target videos. Data construction must maintain their spatiotemporal alignment so that the video model learns to generate the target video while following the proxy. Because action consequences and camera motion are already represented in the proxy video, this interface requires neither separate action labels nor explicit camera-pose annotations, which also makes real-video data easier to incorporate.

\begin{figure*}[t]
  \centering
  \begin{subfigure}[t]{0.495\textwidth}
    \centering
    \begin{minipage}[t]{0.49\linewidth}
      \centering
      \includegraphics[width=\linewidth]{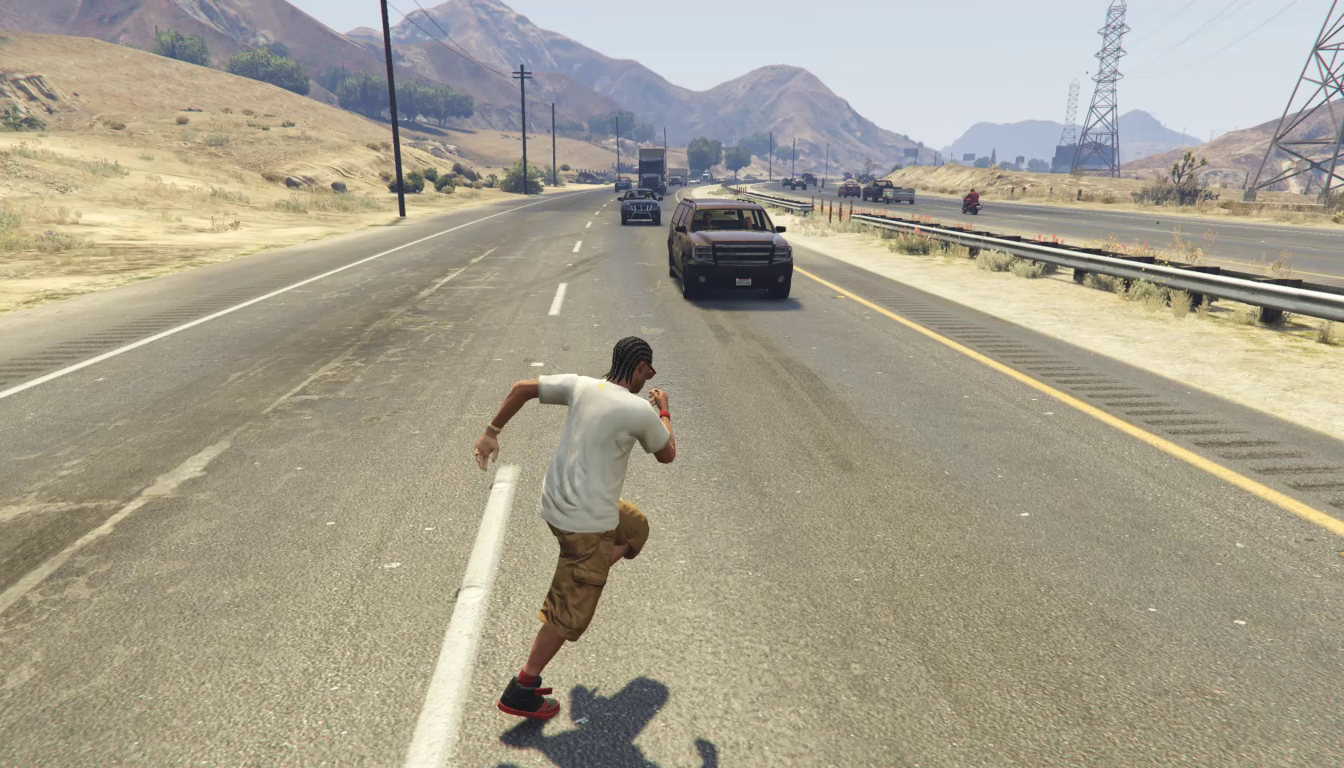}
      \vspace{-2pt}

      {\scriptsize RGB target}
    \end{minipage}\hfill
    \begin{minipage}[t]{0.49\linewidth}
      \centering
      \includegraphics[width=\linewidth]{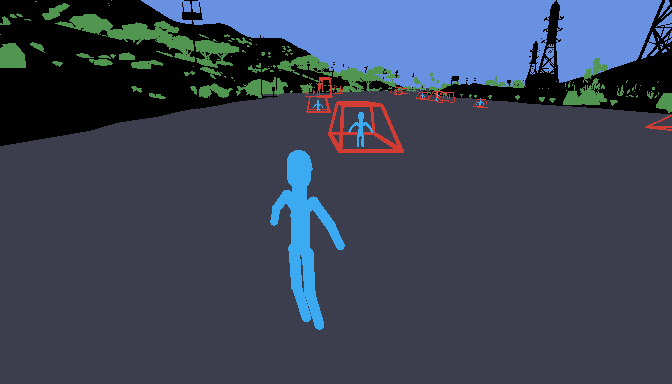}
      \vspace{-2pt}

      {\scriptsize proxy}
    \end{minipage}
    \caption{Game pair data.}
    \label{fig:game_data_proxy}
  \end{subfigure}\hfill
  \begin{subfigure}[t]{0.495\textwidth}
    \centering
    \begin{minipage}[t]{0.49\linewidth}
      \centering
      \includegraphics[width=\linewidth]{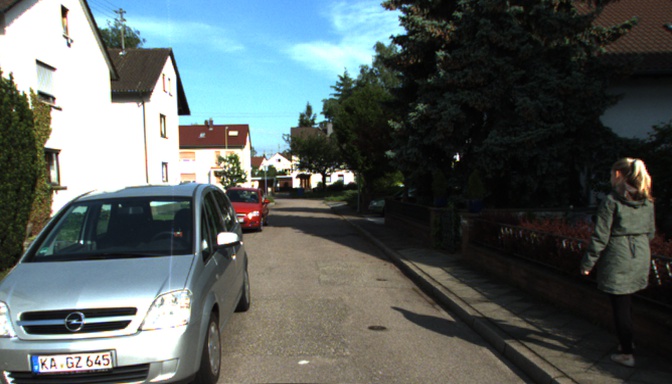}
      \vspace{-2pt}

      {\scriptsize RGB target}
    \end{minipage}\hfill
    \begin{minipage}[t]{0.49\linewidth}
      \centering
      \includegraphics[width=\linewidth]{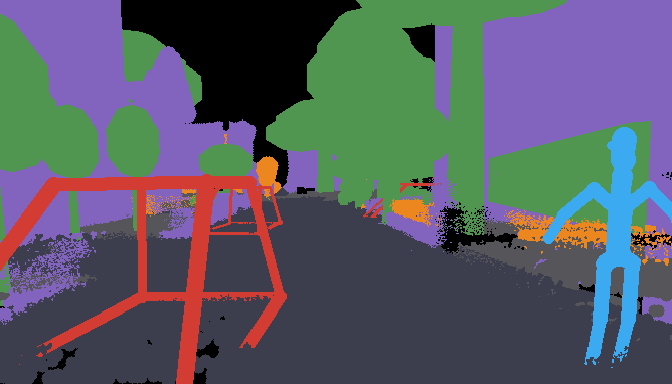}
      \vspace{-2pt}

      {\scriptsize proxy}
    \end{minipage}
    \caption{Real pair data.}
    \label{fig:real_data_proxy}
  \end{subfigure}
  \caption{\textbf{Paired data construction.} Each adjacent RGB--proxy pair is aligned at the same frame. (a) For game data, code reads structured runtime annotations to automatically produce pixel-wise, instance-level proxy annotations and preserve identity across frames. Note that, to ensure that the coding agent can easily reproduce the proxy at inference time, we exclude skeletal pose, detailed 3D models, and other representations that are difficult for coding agents to construct and maintain. Because extraction and compilation are code-based, a coding agent can easily add or revise the retained annotation channels through small, local program changes. (b) For real data, calibrated 3D reconstruction and object annotations compile the aligned proxy offline, preserving scene layout, entity locations, and occlusion without action labels or access to game-engine runtime state.}
  \label{fig:data_pairs}
\end{figure*}

\begingroup
\newcommand{\cwmresultcase}[3]{%
  \begin{minipage}[t]{\textwidth}
    \centering
    \includegraphics[width=\linewidth]{#2}
    \vspace{-3pt}

    {\scriptsize\textbf{Prompt (#1):} #3\par}
  \end{minipage}}

\begin{figure*}[p]
  \centering
  \cwmresultcase{A}{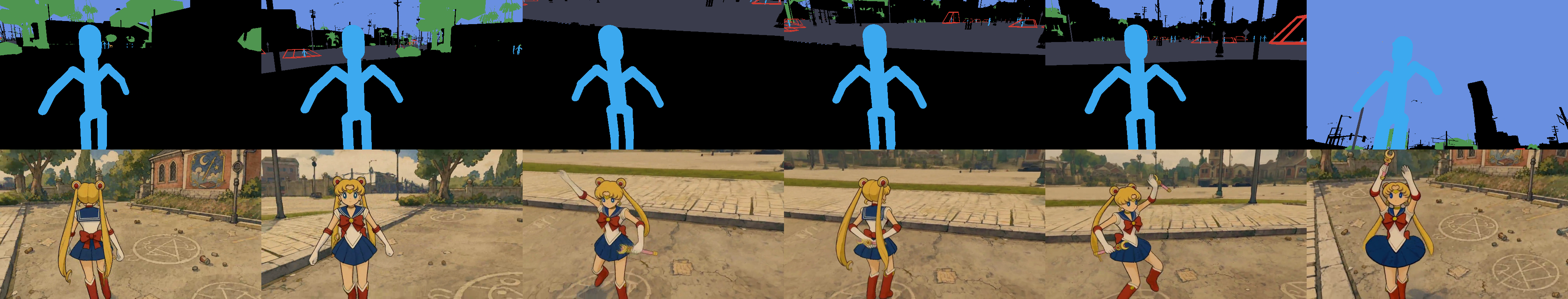}{Sailor Moon performs a lively wand dance in a sunlit rubber-hose wizard academy courtyard.}
  \vspace{2pt}

  \cwmresultcase{B}{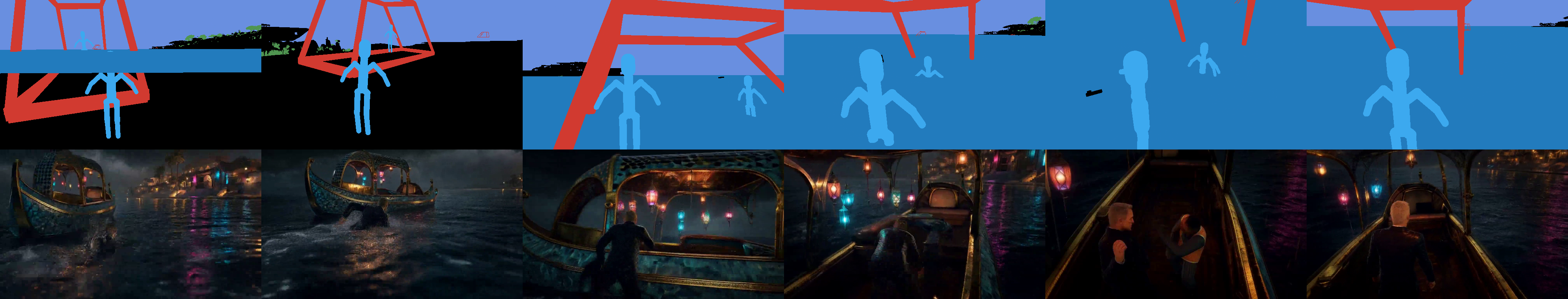}{James Bond swims to a turquoise royal skiff and battles its ferryman in a neon rainy bazaar lagoon.}
  \vspace{2pt}

  \cwmresultcase{C}{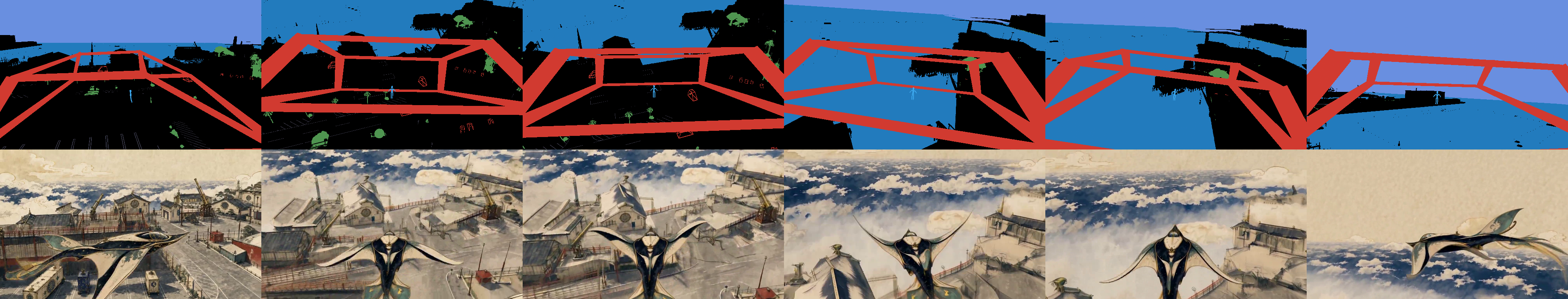}{A celestial manta-skiff glides through sumi-e skies above an ivory observatory and indigo cloud sea.}
  \vspace{2pt}

  \cwmresultcase{D}{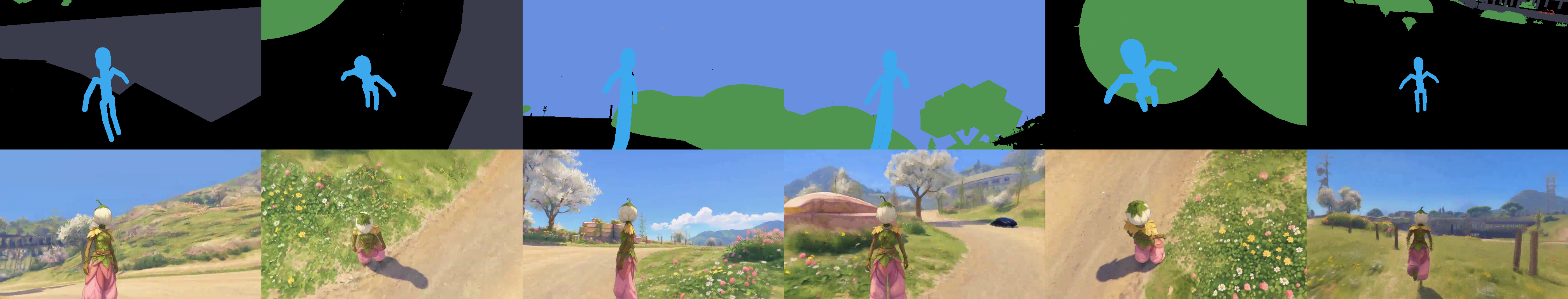}{A dandelion-seed courier runs through a vivid cel-shaded spring meadow past a mossy petal-stone shrine.}
  \vspace{2pt}

  \cwmresultcase{E}{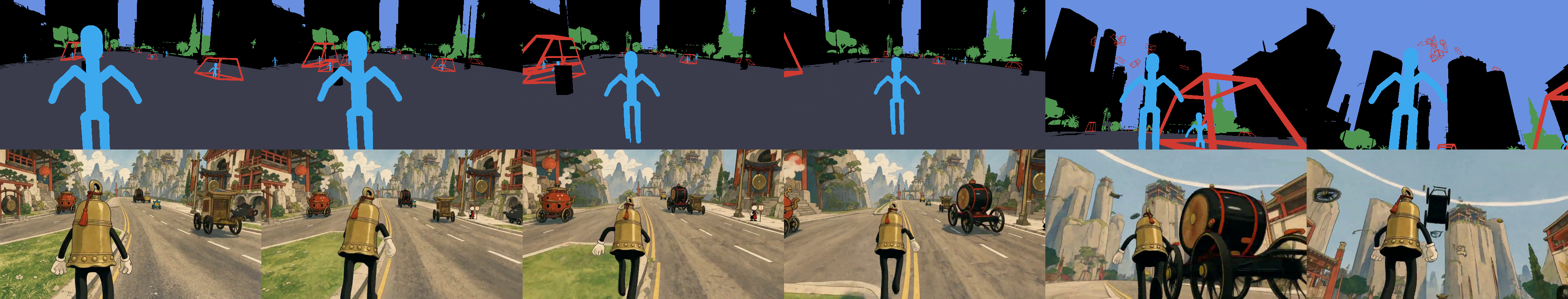}{A rubber-hose brass bell pilgrim runs through a mountain monastery toward a rolling prayer-drum cart.}
  \caption{\textbf{Visual quality results.} Each case contains six temporally ordered frames, with frame-aligned proxies in the upper row and generated RGB observations in the lower row; the corresponding text prompt appears below each pair. Despite using only five hours of GTA V gameplay for fine-tuning, our method exhibits strong generalization across diverse characters, environments, motions, and camera trajectories, preserving rich visual detail while following the spatiotemporal structure specified by the proxy.}
  \label{fig:visual-quality}
\end{figure*}
\endgroup

\paragraph{Game data.}
Games are a particularly suitable source because a target video and its corresponding runtime state can be obtained synchronously from the same execution \citep{kim2020gamegan,alonso2024diamond,valevski2024gamengen}. During gameplay, we record the target video and extract from the runtime only the state required to construct the proxy. We do not retain complete 3D models, textures, or materials; we record camera state, entity positions and orientations, approximate scale, scene layout, and interaction-critical state. We use these variables to programmatically construct a proxy that is rendered along the target video's camera trajectory, forming a condition-target pair for the same gameplay process. The same runtime recording can be recompiled into proxies with different coverage and information granularity without recollecting the target video.

The primitive vocabulary need not be fixed to one shape, granularity, or visual style across all data. We find that contemporary video models can learn correspondences between varied coarse primitives and target visual objects. For complex objects that cannot retain sufficient discriminability through simple primitives, the proxy may use only a bounding box to convey location and coarse extent, while structured text specifies identity and attributes. Any residual ambiguity can be further resolved through identity, appearance, and action semantics in the accompanying text.

Our experiments also indicate that a coding agent can autonomously construct simple but discriminative primitives. For common categories such as people, trees, vehicles, bushes, and ordinary buildings, the agent classifies each identity and composes the corresponding primitives. For structurally complex or uncommon objects, it first decides whether simple primitives preserve sufficient discriminability; otherwise, the system uses a bounding box. The box marks only the identity's location and coarse spatial extent, while text provides category, appearance, attributes, and action. This fallback avoids requiring the coding agent to reconstruct complex geometry while still informing the video model that the object exists at a particular location.

The recorded runtime further provides precise identity correspondence. For each frame of the rendered proxy video, we obtain a pixel-wise ground-truth instance map that identifies the identity associated with each pixel. This mapping binds the appearance, attributes, and action descriptions for every identity in structured text to the corresponding proxy region, yielding a more precise condition. Fig.~\ref{fig:game_data_proxy} shows one such automatically constructed pair.

\paragraph{Real-world data.}
Real videos are another important source. Their value is not only scale, but also high-fidelity appearance, motion, and physical interaction that game assets, animation, simulation, and rendering pipelines cannot fully cover. These priors can move the learned observation model beyond the visual and dynamic ceiling imposed by a particular game pipeline. They are especially important for embodied agents, whose tasks require cues about scale, occlusion, contact, camera egomotion, and the motion and physical response caused by actions. Real-video data can make generated visual environments closer to an agent's deployment distribution, allowing Code World Model to support not only virtual-world generation but also scalable real-world visual environments for training and evaluating agents.

For approaches that depend on separate action or camera conditions \citep{zhang2025matrix_game,sun2025worldplay,wang2026matrix_game3}, turning real videos into training data typically requires action labels or camera trajectories. Actions in unconstrained real-world videos can be defined at different semantic granularities, making annotation ambiguous, while continuous camera motion is also difficult to estimate and label. Our interface requires only a target video, a corresponding proxy video, and a structured prompt. The visual effects of actions and camera motion are represented directly in the proxy video rather than defined as independent labels.

To verify that proxy--observation pairs can be constructed from real-world recordings, we build a geometry-assisted proof of concept on KITTI-360~\cite{kitti360}. Rectified RGB frames serve as target observations, while calibrated camera poses, the accumulated semantic 3D reconstruction, and 3D object annotations are used only during offline data construction to compile the corresponding conditions; none of them is provided to the video model. At each timestamp, we project the static scene geometry into the calibrated camera view to obtain metric depth and depth-derived surface normals. We represent scene elements requiring explicit spatial grounding using the same coarse primitive vocabulary as for game data: capsule skeletons for pedestrians, wireframes for vehicles, low-poly proxies for vegetation, and coarse 3D extents for large buildings. The projected geometry and proxy primitives are rasterized through a shared depth buffer, preserving camera motion, spatial alignment, and occlusion across all conditions. Using the same dataset frame index establishes a one-to-one correspondence between each RGB target and its compiled conditions. Fig.~\ref{fig:real_data_proxy} shows a representative aligned frame, illustrating that proxy conditions can be compiled from an existing real-world 3D reconstruction without access to game-engine runtime state.

\section{Experiments}
\label{sec:experiments}

\subsection{Implementation Details}

\paragraph{Training.}
We fine-tune the official MiniMax-H3 Ref2VA backbone~\citep{minimax2026h3} for proxy-conditioned video generation. Each training sample uses a proxy sequence and structured text as conditions, with the paired RGB video serving as the prediction target. The 157 gameplay takes contain approximately 5.6 hours of source video. We sample 9,420 five-second training clips from these takes at two-second intervals. Each target is a 124-frame video at $1344\times768$ and 24 FPS. Its temporally aligned proxy condition is a 124-frame, $336\times192$ sequence combining fixed-log depth with categorical semantic-ID maps. The multimodal encoder receives a system instruction, a clip-specific text description, and 11 proxy frames sampled at offsets $0,12,24,\ldots,120$. We apply rank-128 LoRA \citep{hu2022lora} throughout all 50 transformer blocks, yielding approximately 596M trainable parameters. Starting from the official Ref2VA checkpoint, we optimize only the video objective on eight NVIDIA H800 GPUs using AdamW with a global batch size of eight, weight decay 0.01, gradient-norm clipping at 1.0, BF16 mixed precision, FlashAttention-3 \citep{shah2024flashattention3}, and gradient checkpointing; the audio loss is disabled. Given our available compute budget, training runs for three epochs, or 3,534 optimizer steps, with a 100-step warmup followed by cosine learning-rate decay from $2\times10^{-5}$ to $1\times10^{-6}$.

\paragraph{Coding-agent inference.}
We use GPT-5.6 Sol as the coding agent. During inference, we provide it with existing game-engine code that supplies basic player controls, collision handling, and a runtime update loop, together with the same proxy primitives used in training-data construction. Building AAA-scale scenes and complex interaction logic entirely from scratch remains difficult for current coding agents. We find, however, that the coding agent can readily use existing AAA game scenes and gameplay logic as templates, combining, extending, and rewriting them to construct each target world. The resulting world is executable and player-controllable, while representing visually relevant state with coarse proxy geometry rather than refined 3D assets. We run this world under user control, record the resulting proxy video, and provide that sequence as a frame-wise condition to the downstream video model. Given the rapid progress of coding agents, we expect them to increasingly construct complex world interactions and scene layouts from scratch.

\paragraph{Video-model inference.}
We use the checkpoint at step 3,534 and sample each video with explicit Euler for 20 steps. For the qualitative examples, GPT Image~2 generates a $1536\times864$ first-frame appearance anchor conditioned on the first proxy frame and a text prompt written by the coding agent. The coding-agent-written image prompt instructs the model to preserve the proxy-specified entity count, approximate positions, relative scale, object orientations, coarse layout, camera height, and framing, while realizing the identities, materials, environment, and overall visual style described in text. The generated image is resized to $1344\times768$, initializes target latent slot 0, and is supplied to both the multimodal encoder and the Ref2VA appearance-reference branch. The complete proxy sequence provides frame-aligned control over camera motion, entity motion, and scene layout, while text specifies appearance, action semantics, and details not represented by the proxy. Each inference call generates a 124-frame video at $1344\times768$ and 24 FPS. Inference is video-only and uses FlashAttention-3 on H800 GPUs.

For long-video inference, we run the model over overlapping 124-frame windows with a 34-frame overlap, so each new window advances the video by 90 frames. The first window is initialized with the GPT Image~2 appearance anchor described above. For every subsequent window, the final 34 RGB frames from the preceding window provide continuation context, while the same appearance anchor is again supplied to both the multimodal encoder and the Ref2VA appearance-reference branch. The overlapping RGB context maintains local temporal continuity, whereas reusing the anchor helps preserve global identity and appearance across windows. Each call generates a complete 124-frame window; during stitching, we retain all frames from the first window, discard the regenerated 34-frame overlap from each later window, and append its remaining 90 frames. Frame-aligned proxy conditions are provided throughout the sequence. Their $336\times192$ spatial resolution is one quarter of the $1344\times768$ output resolution along each dimension, corresponding to one sixteenth as many pixels. Consequently, even frame-wise proxy conditioning introduces few additional visual tokens and adds negligible inference overhead relative to full-resolution video generation.

\subsection{Visual Quality Results}

We present visual quality results in Fig.~\ref{fig:visual-quality}. Each video is conditioned on a proxy sequence recorded from a simple interactive world constructed by the coding agent, together with a text prompt written by the same agent. We find that, even after LoRA adaptation on only a small amount of paired training data, the video model closely follows proxy-specified entity positions and motion, scene layout, and camera trajectories.

Our qualitative comparison videos on the project page show that proxy conditioning provides more precise and responsive control over character motion, actions, and camera movement than action- or camera-conditioned video world models \citep{sun2025worldplay,wang2026matrix_game3,gao2026lingbot2}. By specifying frame-wise poses and trajectories directly, the proxy controls individual character actions and camera operations at the same granularity as a modern 3D game. We refer readers to these demos for the complete temporal comparisons. To isolate control quality, the comparisons do not consider inference latency.

\section{Limitations and Conclusions}
\label{sec:limitations-conclusions}

\noindent\textbf{Limitations.} Our current system has two primary limitations. First, our experiments are limited by available compute. The training scale therefore remains small, and the resulting generation quality is still limited. We also do not implement autoregressive real-time generation. Second, despite their rapid progress, current coding agents still struggle to implement highly complex game mechanisms reliably from scratch. Our prototype therefore does not demonstrate autonomous construction of a complete open-world game or simulator. We expect future coding agents to design, implement, test, and maintain increasingly complex world mechanisms, but reliably achieving this capability remains future work.

\noindent\textbf{Conclusions.} We introduced \textbf{Code World Model}, which separates world evolution from visual realization: a coding agent uses executable code to maintain persistent world state, while a video model generates visual observations. A \textbf{proxy} connects these components through frame-wise spatiotemporal conditions, and our data pipelines construct aligned proxy--observation pairs for training. After adaptation on paired gameplay data, qualitative evaluations using proxies recorded from simple coding-agent-built worlds show that the video model can follow these proxy conditions while preserving visual detail and dynamics, suggesting a promising path toward open-ended world models.

\clearpage
{
  \small
  \bibliographystyle{ieeenat_fullname}
  \bibliography{references}
}

\clearpage
\appendix
\begin{center}
  {\Large\bfseries Appendix}
\end{center}
\section{Text Prompt Examples}

\begingroup
\fvset{fontsize=\footnotesize,breaklines=true,breakanywhere=true,
  frame=single,framesep=2mm}

\begin{Verbatim}
Appearance and scene:

The protagonist is the solitary reed-masked rainkeeper, an adult humanlike valley guardian with a sturdy, upright build. A softly weathered celadon mask covers the entire face, forming a smooth brow, straight nose, closed lips, and calm sculpted cheeks above a firm jaw. Narrow dark eye openings remain shadowed beneath a broad, low conical rain hat woven from wet black-brown reeds. The celadon mask, shadowed eyes, humanlike proportions, and reed-covered silhouette are stable identity features. A dense mantle of dark woven river reeds spreads across the shoulders and hangs in layered, uneven lengths past the hips, with individual rain-heavy stalks, cross-ties, and ragged tips clearly visible. Beneath it, a deep indigo raincloth wraps the neck and chest, overlaps into a long weather-darkened tunic, and gathers beneath a thick knotted rope belt. Narrow wrapped sleeves end in dark practical gloves. Loose charcoal trousers descend into mud-dark boots, although the initial lower edge crops the figure at the lower shins and hides both feet and ground contact. Moisture deepens the indigo cloth, gives the reed mantle a subdued sheen, and gathers as restrained silver highlights along the hat brim, mask, gloves, and layered hems. The solitary reed-masked rainkeeper begins at lower center-left, shown nearly in right-facing profile with the whole body oriented toward screen-right. The posture is neutral and balanced, the shoulders remain level beneath the broad mantle, the arms hang naturally beside the torso, and the masked head follows the body's forward direction.

No other character, creature, mount, vessel, or moving identity occupies the sanctuary. The solitary reed-masked rainkeeper remains the only salient living figure throughout the visible landscape.

The rainkeeper stands upon an elevated, walkable stone spillway crossing a vast monsoon-washed terraced valley. Nearest the camera, broad charcoal flagstones form an open, unobstructed surface with a damp diagonal perspective. Pale channels inset between the stones carry thin, fast-moving ribbons of rainwater, their broken silver reflections leading toward screen-right and into the distance. A low moss-dark parapet extends across the middle distance beyond the rainkeeper. Past it, immense stepped rice terraces descend through flooded emerald levels, slate retaining walls, rain cisterns, and clustered rooflike irrigation shrines. Water spills in narrow sheets from ledge to ledge, softening the masonry with mist and wet reflected light. Tall bamboo rain gauges rise among the terraces, including a prominent clustered gauge near center-right, while taut black irrigation ropes cross the valley between their upper fittings. Layered mountain silhouettes recede through humid aerial haze beneath an open storm-dark sky. Heavy monsoon clouds range from deep charcoal to cool blue-gray, with veils of distant rainfall and a weak break of dramatic side light touching wet stone, celadon, and water. Deep viridian vegetation, wet umber masonry, muted indigo cloth, and restrained silver reflections remain unified by moody classical oil-painting realism, visible disciplined brushwork, rich glazing, and atmospheric depth. The camera begins in a wide horizontal view near the rainkeeper's upper-torso height, viewing the figure from the left side with the lower shins cropped; it then arcs smoothly behind the stationary rainkeeper, holds a broad rear view across the spillway, and returns toward the initial side profile near the end.

Protagonist action timeline:

0.00-0.75s: The solitary reed-masked rainkeeper holds the exact opening stance at lower center-left, upright in near right-facing profile with the masked face and torso directed toward screen-right. Both arms remain relaxed, the gloved hands hang beside the thighs, and the broad reed mantle settles heavily over the shoulders while the camera begins moving from the rainkeeper's left side toward a rear three-quarter view.; 0.75-1.50s: The solitary reed-masked rainkeeper remains planted without stepping as the camera continues its smooth arc behind the body. The visible profile narrows, the back of the conical reed hat and layered mantle becomes dominant, and the figure shifts toward the lower center of the composition while still facing along the same open length of spillway.; 1.50-3.75s: The solitary reed-masked rainkeeper stands motionless in a centered rear view, maintaining level shoulders, a straight torso, softly bent elbows, and evenly lowered hands. Only restrained breathing lifts the layered raincloth, loose reed tips stir under the rainfall, and water glides continuously through the pale channels as the camera holds nearly steady behind the figure.; 3.75-4.50s: The solitary reed-masked rainkeeper preserves the same footing and world-facing orientation while the camera resumes its lateral arc from the rear toward the figure's left side. The mantle's rear layers gradually give way to the celadon mask's returning profile, and the rainkeeper drifts visually from lower center toward lower center-left as the terraced valley slides across the background.; 4.50-5.17s: The solitary reed-masked rainkeeper finishes in a near right-facing side profile closely matching the opening orientation, with the chin level, arms lowered, gloves still, and weight evenly supported. The camera settles into a broad side view as rain beads on the hat brim and mantle, silver water continues coursing through the inset channels, and the solitary figure remains upright against the storm-dark terraces.

Other visible actors:

0.00-5.17s, no other visible actors: The spillway, parapet, stepped terraces, irrigation shrines, bamboo rain gauges, distant mountains, and storm sky remain unoccupied by any other people, creatures, mounts, vessels, or independently moving identities throughout.
\end{Verbatim}

\noindent\textbf{Example 1: The Rainkeeper's Steps.} The subject remains stationary while the camera arcs from a side view to a rear view and returns, isolating camera following from entity translation.

\medskip
\begin{Verbatim}
Appearance and scene:

The protagonist is the lone candle-headed marionette courier, a lanky carved wooden figure with a narrow torso and long jointed limbs. The courier's body is built from angular dark-teal timber segments joined at the shoulders, elbows, wrists, hips, knees, and ankles by visible polygonal pivots. Blocky wooden hands terminate the slim forearms, and wedge-shaped feet support the elongated legs. A small glowing candle-flame head is enclosed inside a simple angular glass lantern, with warm amber light shining through its broad faceted panes and a dark metal cap forming the lantern's pointed crown. The steady flame, lantern head, dark-teal timber body, narrow marionette proportions, and articulated wooden limbs are stable identity features. A short plum-colored tailcoat made from broad polygon planes wraps the torso, rises into a stiff open collar beneath the lantern, and divides into pointed tails behind the hips. A narrow dark belt with two muted brass studs closes the coat at the waist. Cool lavender moonlight gives the coat's outward planes dusty violet edges while the lantern casts restrained amber light over the collar and upper chest.

At the opening instant, the courier stands in the lower center-left from a rear three-quarter view, facing screen-right and slightly uphill. The right foot is planted ahead on a higher slate facet, the left leg trails diagonally downhill, the torso remains nearly upright, and both arms hang loosely with slight bends at their wooden joints. The broad hillside thoroughfare fills the foreground and extends unobstructed toward the upper right. Its traversable surface consists of irregular charcoal slate and muted ochre cobbles assembled from crisp triangular facets, with shallow ridges and color changes defining the steep ascent rather than rails or walls. Downhill behind the courier, the terrain opens into a deep violet ravine containing a narrow pale stone road that winds among the cliffs and climbs toward the far heights.

Layered polygonal cliff masses surround the ravine from middle distance to the horizon. Sparse crooked Halloween-town buildings are fused into the darker slopes, with steep roofs, shuttered facades, narrow towers, and small amber-lit windows. Static geometric pumpkins sit beside a few distant houses and among the lower rock ledges, while pointed moss-green trees break up the charcoal and dusty-violet architecture. The entire town uses crisp flat-shaded geometry, simplified silhouettes, matte surfaces, hard polygon edges, and visibly triangular planes. Cool lavender moonlight keeps the open route, ravine, and distant road clearly readable, while the lantern and scattered windows provide restrained warm accents. The trailing camera begins close behind the courier near upper-torso height, holding the exact rear three-quarter layout before following the courier toward the crest and then rising, rolling, and widening with the ensuing fall to reveal more of the ravine, cliffs, winding roads, and hillside town.

Protagonist action timeline:

0.00-0.75s: The candle-headed marionette courier continues directly from the opening walking pose, settling weight through the forward right foot while the left heel lifts from the lower slate. The narrow torso inclines slightly uphill, the right arm drifts behind the hip, and the left arm begins a small forward swing as the lantern remains upright.; 0.75-1.50s: The courier brings the left leg through and places the left foot farther uphill, completing one measured stride across the uneven facets. The hips advance toward screen-right, the coat tails separate slightly over the moving thighs, and the wooden elbows maintain a loose alternating swing.; 1.50-2.25s: The right foot lifts, passes the planted left leg, and reaches onto another ochre-edged slate plane. The courier's weight rolls forward through the narrow torso, the lantern flame stays enclosed and steady, and the camera trails at nearly the same distance while the convex crest grows nearer.; 2.25-3.00s: The courier takes another short uphill step with the left foot and moves onto the rounded high part of the thoroughfare. Both knees flex to accommodate the sloping surface, the shoulders remain angled away from the camera, and the pointed plum coat tails sway once behind the hips.; 3.00-3.50s: Near the crest, the courier's balanced gait breaks abruptly as the upper body pitches outward toward the open ravine. The forward foot loses firm support, both arms rise and spread from the shoulders, the trailing leg straightens behind, and the lantern head tilts with the torso while the body passes beyond the descending edge.; 3.50-4.00s: Both wedge-shaped feet leave the slate and the courier falls freely into open space. The body extends nearly sideways above the ravine, one arm reaching ahead and the other trailing, as the hips roll and the long legs separate into a loose scissor shape. The camera swings outward and downward to retain the full wooden figure.; 4.00-4.50s: The courier descends rapidly while rotating forward, bringing the lantern head lower than the hips. The shoulders roll toward the cliffs below, both elbows bend under the tumble, and the plum coat tails flare away from the thighs as the winding pale road and dark slopes spread across the background.; 4.50-4.75s: The forward rotation carries the legs upward behind the torso until the courier becomes briefly steep and head-down. The knees bend, the feet sweep through the upper part of the view, and the lantern's amber flame remains visible through the turning glass enclosure.; 4.75-5.17s: The courier continues falling and rolls from the head-down posture into a broad sideways tumble above the ravine. The long legs fold and then begin extending apart, the arms remain unevenly outstretched, and the coat tails flutter close to the rotating hips while the widening camera follows from above and behind without interrupting the descent.

Other visible actors:

No other visible actors appear from 0.00-5.17s. The crooked hillside buildings, geometric pumpkins, pointed trees, amber windows, layered cliffs, and winding pale roads remain static environmental elements while the lone candle-headed marionette courier crosses the crest and falls through the open ravine.
\end{Verbatim}

\noindent\textbf{Example 2: Hollowfacet Hill.} The subject walks uphill, loses balance, and enters a continuous fall, providing a prompt with several temporally ordered motion states.

\medskip
\begin{Verbatim}
Appearance and scene:

The protagonist is an adult shoebill courier with a tall, slender avian build. It stands in the lower center-left, seen from behind with its body facing downhill and slightly toward screen-right. Dense slate-blue plumage covers its narrow neck, broad back, folded wings, and tapered tail, with layered flight feathers forming crisp overlapping rows. Its large head carries a short swept crest, a small pale eye visible in profile, and an immense weathered gray bill with tan mottling and a deep hooked tip. The head is turned slightly toward screen-right while the shoulders and torso remain aligned with the descending road. Its wings hang naturally beside the body, their long outer feathers reaching toward the thighs. Dark stiltlike legs descend from beneath the tail, but the feet and ground contact are not visible beyond the lower edge. The slate-blue feather layers, massive mottled bill, compact crest, long dark legs, and upright shoebill silhouette are stable identity features. A practical woven reed pannier rests across the courier's lower back. Thick braided straps pass over both shoulders, and the rounded basket is packed with colorful seedpods, dark perforated pods, small gourds, waxy berries, and curled dried fruit in muted ochre, berry red, plum, and moss green. The pannier rides snugly against the feathers, with its fibrous rim and irregular weave visibly worn by use.

A broad lane of pale packed clay descends from the courier into an unstaffed biological market. Shell fragments, fine gravel, shallow ruts, exposed roots, scattered twigs, and damp darker patches break up the firm traversable surface. The road remains open directly ahead, narrowing gently between low displays before reaching the distant settlement. On screen-left, a mossy earthen wall rises close beside the courier beneath enormous shelf fungi. Their broad ochre caps project outward as layered awnings, with fibrous undersides, rain-darkened edges, and small trailing roots. Pitcher plants and shallow woven trays beneath them contain green fruit, dark nuts, reddish tubers, seed clusters, and folded leaves. A tall clay vessel and a short cut-stem container sit near the road edge without obstructing the lane.

On screen-right, a raised bank supports further shelf-fungus awnings and woven root counters. Baskets and low trays hold orderly piles of orange fruit, red berries, pale gourds, spices, and glossy green produce. Tall pitcher plants rise at the nearest corner, their ribbed green and burgundy cups opening upward beside the road. Short wooden posts support the counters, while small jars, knotted roots, moss, and damp leaf litter fill the recesses beneath them. A colossal kapok trunk climbs beyond the bank at the upper screen-right. Farther downhill, additional fungus-roofed displays cluster on both sides while keeping the center of the road clear.

Immense weathered termite-spire storehouses dominate the distant settlement. Their eroded clay surfaces rise in irregular pinnacles, hollow apertures, buttresses, and broken crownlike ridges, interspersed with hollow kapok towers and dense green trees. Humid haze softens the farthest structures without obscuring their vertical silhouettes. Restrained moss green, clay ochre, berry red, slate gray, and damp earth tones remain natural throughout. Soft overcast midday light reveals accurate feather barbs, fungal fibers, woven reeds, pitted clay, and soil texture without theatrical glow. The trailing camera begins close behind the courier at upper-back height with subtle telephoto compression, holding the courier large in the foreground while the descending market road and termite-spire settlement remain fully legible ahead.

Protagonist action timeline:

0.00-0.75s: The slate-blue shoebill courier holds the exact opening stance near the road center-left, body directed downhill and slightly toward screen-right, head angled toward screen-right, folded wings hanging beside the pannier, and feet remaining outside the visible area. Slow breathing gently lifts the upper back and shifts the woven basket against the feather layers.; 0.75-1.50s: It makes a restrained idle weight shift through its long legs without advancing, allowing the torso to drift slightly toward screen-left while the head remains elevated and the heavy bill stays directed diagonally toward the open road. The seedpods settle together with a small delayed movement inside the pannier.; 1.50-2.25s: The courier eases back toward a centered balance, lowers the wing on screen-right a little away from its flank, and then lets the feathers fall back into their layered resting position. Its neck straightens subtly while the bill retains the same general downhill orientation.; 2.25-3.00s: It rotates its head a small amount farther toward screen-right as if surveying the nearest produce counters, while its shoulders, torso, and planted lower-body position remain aligned with the lane. The crest feathers and loose neck plumage respond with a faint natural tremor before settling.; 3.00-3.75s: The head returns gradually toward the road ahead, led by the bill and followed by a slight neck roll. The courier shifts pressure toward its opposite leg, causing the tail feathers and pannier to sway together by a small amount without any downhill travel.; 3.75-4.50s: It completes the gentle sway and resumes an upright, balanced posture. Both folded wings hang close to the body, the shoulder straps remain taut across the back, and the basket stops moving after one soft pendular correction.; 4.50-5.17s: The slate-blue shoebill courier remains stationary and continues looking downhill, with only quiet breathing, a minute head adjustment, and slight feather settling visible. Its body orientation, foreground scale, relationship to the open lane, and distance from the nearest stalls remain unchanged through the end.

Other visible actors:

0.00-5.17s, the rooted biological market: The giant shelf-fungus awnings, pitcher-plant bins, woven root counters, arranged produce, immense termite-spire storehouses, and hollow kapok towers remain fixed around the lone courier, with only slight movement in small leaves and exposed fungal fringes under the humid overcast air.
\end{Verbatim}

\noindent\textbf{Example 3: The Mycelial Provision Road.} The subject remains in place while performing subtle idle motion, testing fine motion and scene preservation without large displacement.

\endgroup

\end{document}